\documentclass[conference]{IEEEtran}
\usepackage{amssymb}
\usepackage{graphicx}

\ifCLASSINFOpdf
\else
\fi
\usepackage{algorithm}
\usepackage{algorithmic}
\usepackage[tight,footnotesize]{subfigure}
\begin{document}
%
\title{Classifiers With a Reject Option for Early Time-Series Classification}

\author{\IEEEauthorblockN{Nima Hatami}
\IEEEauthorblockA{Department of Ophtalmology\\
University of California \\
9415 Campus Point Drive, La Jolla, CA  92093, USA \\
Email: nhatami@ucsd.edu}
\and
\IEEEauthorblockN{Camelia Chira}
\IEEEauthorblockA{Department of Computer Science \\
Babes-Bolyai University \\
Cluj-Napoca 400084, 1 Kogalniceanu, Romania\\
Email: cchira@cs.ubbcluj.ro}}

\maketitle

\begin{abstract}
Early classification of time-series data in a dynamic environment is a
challenging problem of great importance in signal processing.
This paper proposes a classifier architecture with a reject option
capable of online decision making without the need to wait for the
entire time series signal to be present. The main idea is to classify
an odor/gas signal with an acceptable accuracy as early as possible. 
Instead of using posterior probability of a classifier, the proposed
method uses the "agreement" of an ensemble to decide whether to accept
or reject the candidate label. 
The introduced algorithm is applied to the bio-chemistry problem of
odor classification to build a novel Electronic-Nose called
\emph{Forefront-Nose}. 
Experimental results on wind tunnel test-bed facility confirms the robustness of the forefront-nose compared to the standard classifiers from both earliness and recognition perspectives.  

\end{abstract}


%
\IEEEpeerreviewmaketitle

\section{Introduction}

An electronic nose is a device intended to detect and recognize different odors or flavor types.
Over the last decade, electronic sensing also known as
e-sensing technologies have undergone important developments
from a technical and commercial point of view \cite{Hines1999,
  Arshak2004, Gutierrez2002, James2005}. E-sensing refers to the
capability of reproducing human senses using sensor arrays, pattern
recognition and machine learning techniques. In international space
stations for instance, E-Noses are designed to help reducing the risks
associated with crewmember exposure to toxic or dangerous
chemicals. Conformity of raw materials and final products, detection
of contamination, spoilage, adulteration and monitoring of storage
conditions widely use E-Nose technology at line quality
control. Identification of volatile organic compounds in air, water
and soil samples for environmental protection and monitoring purposes
represent only a few of the application areas where odor/flavor recognition plays a key role.   

From machine learning prospective, any gas/odor signal produced by a
sensor can be considered as a sequence of pairs \emph{(timestamp,
  value)} and these data values are ordered in timestamp ascending
order. Similarly, the odor recognition task in e-sensing can be viewed
as a time series classification problem in pattern recognition. 
Overviewing the gas signal processing literature, there are three
major types of approaches to deal with the odor classification task \cite{Hines1999, Arshak2004, Gutierrez2002, James2005}. The first category is feature-based classification, which transforms a signal sequence into a feature vector and then applies conventional classifiers. Feature selection plays an important role in this kind of methods.
The second category is sequence distance-based classification. The distance function which measures the similarity between sequences affects the recognition rate significantly. 
The third category is model-based approaches such as hidden markov
models (HMM) and other statistical models. It is worth noting that the
traditional methods for odor identification (and time series
classification, in general) are mostly \emph{whole-sequence based}
i.e. they make an a-priori assumption that all the elements of a time series signal to be classified are observed in advance.

With the increasing availability of temporal data at a large scale, there is a growing demand for early classification of an ongoing sequence as early as possible, preferably after
only few of its preceding elements occurred. For example, many air
quality problems caused by toxic chemical leaks or spills have
occurred on numerous international space stations and space shuttle
flights. In most of these events, the problem chemical(s) were either
never identified or were identified only after the crew had been
exposed to it. These represent significant health and safety risks to
the crewmembers. Real-time operating E-Nose is designed to help
reducing risks associated with crewmember exposure to toxic or
dangerous chemicals. Besides the safety issues, dealing with large
memory requirement and computational reduction is another motivation
for the investigation of real-time classification. However, constructing classifiers that are capable of early prediction is far from trivial. In the traditional sequence/time series classifiers, the optimization goal is often set to maximize recognition rate. However, in early classification, the goal is to optimize the \emph{earliness} as
long as the classification accuracy is satisfactory.

This paper proposes a model based on a set of serially located
classifiers with a reject option to address the early classification
of time-series. The first classifier makes a decision about the type
of an incoming gas/odor based on a small portion of the signal
available or rejects the sample leaving it to the next classifier. The
decision making of the next classifier is based on the new
portion of the time-series signal which could have been not available
for the first classifier. The second classifier assigns a confident
label to the sample or passes it to the next classifier in the
set. This online process continues in this manner in an iterative manner. It is worth noting that the classification cost
of a time-series signal by the second classifier is higher compared to
that of the first one, since earliness of a decision making is
important in the considered problem. At each step, the decision making
will be made based on a portion of the available signal or it will be postponed to the next available patch of signal to be handled by the next classifier. This process can be repeated  until a classifier
is confident enough about the label or the cost of postponing the
decision is too high.

Computational experiments focus on the bio-chemistry problem of
odor classification. A novel Electronic-Nose called
\emph{Forefront-Nose} is created based on the proposed model for early
classification of time-series. The results obtained on the wind tunnel
test-bed facility indicate the robustness of the Forefront-Nose
compared to the standard classifiers from both earliness and recognition perspectives.  

The structure of this paper is as follows: section 2 provides a brief
review of recent machine learning approaches in E-Nose systems
and relevant time-series classification approaches
for the E-Nose application; section 3 focuses on the importance of
early identification of time series signals; section 4 presents
first the theoretical framework of a classifier with a reject option and then introduces the proposed
Forefront-Nose for early recognition of odor types; section 5
presents the computational experiments starting from data acquisition
and measurement procedure, Forefront-Nose hardware description to
numerical analyses and comparisons; and section 6 contains the
conclusions of the paper.

\section{Recent time-series classification approaches in E-Nose systems}

In this section, some recent important progresses in time-series classification of E-Nose systems are discussed. 

Wang et al \cite{Wang2009} proposed an approach based on the relevance vector machines (RVM). The electronic nose data are first converted into principal components using the principal component analysis (PCA) method and then directly sent as inputs to a RVM classifier. The experiments are performed using different combinations of original coffee data and compared to the support vector machines (SVM), the RVM method can provide similar classification accuracy with fewer kernel functions.

In \cite{Fu2007}, an olfactory neural network called the KIII model is
introduced. The distributed open-ended structure of KIII model is
compatible of any dimension of input vector but the more needed
running time is costly. As an extension to their previous work, Fu et
al \cite{Fu2012} study the relationship between classification
performance of the KIII model and the outer data factor, i.e.,
the dimension of input feature vector, as well as the inner
structure factor, i.e., the amount of its corresponding
parallel channels. The PCA technique was applied for feature extraction and dimension reduction. Two data sets of three classes of wine derived from different cultivars and five classes of green tea derived from five different provinces of China were used for experiments.

SVMs are applied for olfactory signal recognition in
\cite{Distante2003}. Different types of kernels are experimented on
two binary datasets and the results are compared to Neural Network
algorithms i.e. Radial Basis Functions and the error backpropagation
algorithm. At the same research line, recently, Vembu et al
\cite{Vembu2012} investigates the benefits of using time-series
features and kernels for two tasks of odor discrimination and
localization. In this work, specialized features and kernels for
time-series data are designed and used in conjunction with SVMs. To
take full advantage of the temporal information in the data, Vembu et
al applied time-series models such as autoregressive models and linear
dynamical systems to extract features from the input signal, and with
designing similarity measures (kernels) for time series. The methods
are validated on an extensive real data set collected in the wind
tunnel test-bed facility (used also for the experimental section of
the current paper). 

Trincavelli et al. \cite{Trincavelli2010} proposed an electronic nose
for identification of bacteria which is present in circulating blood
and causes "Sepsis", also known as blood poisoning of septicaemia. For
classification of each bacteria, during a measurement, the sampling
cycle is repeated ten times. SVM is applied independently to the ten
samples, and therefore, ten estimates $P(C \mid x_{i})$ are obtained,
where $i$ is the number of the sample for the same bacteria. This
estimation is ensembled across ten consecutive responses of the same
sample in order to make the classification more reliable. 
If a mean with significant superior confidence interval for a class is
disjoint and above all the others then classification is performed
(assigning the sequence of samples to that class); otherwise, a
rejection is declared. The method uses features, which capture the
static response (the difference between the value that the sensor has
at the end of the sampling phase minus the baseline value) and dynamic
(the average of the derivative of the sensor during the first 3s of
exposition of the array to the headspace) properties of the signal
from the gas sensor array. The main drawback of such system is
re-measuring the same observation for many times. While in medical
diagnoses it might make sense, this can be too costly and time
consuming for online applications.  

An on-line and portable E-Nose is designed and developed in  \cite{Ozmen2009} for a qualitative discrimination among different high-concentration gas samples. The main refinement refers to using the PCA method rather than a classification, building a system that can measure the quantitative and/or qualitative gas properties with a metric. It is observed that different concentration amounts of a specie follow a certain route in the PCA plots, even if they are in different experiments.


\section{Early recognition of time-series signals} 

For temporal symbolic sequences and time-series, the values of a
sequence are received in time stamp ascending order. Sometimes,
monitoring and classifying sequences as early as possible is
desired. As mentioned above, most of the existing methods assume that the data resides in main memory and is processed offline. However recent advances in sensor technologies require resource-efficient algorithms that can be implemented directly on the sensors as real-time algorithms.

To the best of our knowledge, Diez et al. \cite{Diez2001} first mentioned the concept of early recognition of time-series. They describe a time series by some relative literals, such as "crease" and "stay", and some region literals, such as "always" and "sometimes" over some intervals. Each literal and its associated position are viewed as a base classifier. Adaboost \cite{Freund1997} is used to ensemble the base classifiers. The ensemble classifier is capable of making predictions on incomplete data by viewing unavailable suffixes of sequences as missing features.

Anibal et al. \cite{Anibal2005} apply a case-based reasoning method to classify time series and monitor the system failure in a simulated dynamic system. The KNN classifier is used to classify incomplete time series using various distances, such as euclidean distance and dynamic time warping (DTW) distance. The simulation studies show that, by using case-based reasoning, the most important increase of classification accuracy occurs on the prefixes through thirty to fifty percent of the full length.

Although in \cite{Anibal2005, Diez2001}, the importance of early
recognition on time series is identified and some encouraging results
are shown, the study only treats early recognition as a problem of
classifying prefixes of sequences. Xing et al. \cite{Xing2008} point
out that the challenge of early recognition is to study the trade-off
between the earliness and the accuracy of classification. The methods
proposed in \cite{Anibal2005, Diez2001} only focus on making
predictions based on partial information but do not address the issue
of how to select the shortest prefix to provide a reliable prediction. This makes the result of early classification not easily used for further actions.

Xing et al. \cite{Xing2008} formulate the early recognition problem as
classifying sequences as early as possible while maintaining an
expected accuracy. A feature based method is proposed for early
classification on temporal symbolic sequences. First, a set of
features that are frequent, distinctive and early is selected and then
an association rule classifier or a decision tree classifier using
those features is built. In the classification step, an incoming sequence is matched with all rules or branches simultaneously until on a prefix, a matching is found and the sequence is classified. In this way, a sequence is classified immediately once the user expected accuracy is achieved. The method proposed in \cite{Xing2008} shows some successes in handling symbolic sequences by achieving competitive accuracies using only less than half of the length of the full sequences.

One disadvantage of the method from \cite{Xing2008} is that it cannot handle numeric time series well. Since numeric time series need to be discretized online, the information loss makes some distinctive features not easy to capture. Xing et al. \cite{Xing2009} propose an early classifier for numeric time series by utilizing instance based learning. The method learns a minimal prediction length (MPL) for each time series in the training dataset through clustering and uses MPLs to guide early classification. As shown in section 2, 1NN classifier with Euclidean distance is a highly accurate classifier for time series classification. One interesting property of the method in \cite{Xing2009} is that without requiring a user expected accuracy, the classifier can achieve early classification while maintaining roughly the same accuracy as a 1NN classifier using full length time series.

Early recognition has never been investigated for E-Nose systems,
although it has an essential importance for such an application. In
fact, early classification of gas/odor time-series signals can benefit
an E-Nose from two different perspectives: i) for industrial
applications such as food industry, on-line odor/gas identification
eliminates the need of waiting for entire time-series signal to be available; ii) and
early detection of toxic treats caused by presence of poisonous
gases/chemicals such as those in space shuttles.

Despite the progress in E-Nose development, there are two main issues
with the state-of-the art systems. Almost all classification
techniques will take the entire signal, record it in memory and
process it in order to make a decision. In other words, they function
offline and this requires large amount of time which may cause
problems in real-world applications. Therefore, the \emph{"earliness"}
of the decision in E-Nose applications has enormous importance. The
second issue is the \emph{reliability} of an E-Nose output. Once a
label is assigned into a gas sample, the next question is "how much
does the user should trust this?". A far more user friendly approach
would be if the automatic system confirms itself the reliability of the label.

\section{Proposed Forefront-Nose}
This section presents the proposed method for early and accurate
identification of odor types from time-series using a set of classifiers with a reject options (CWRO). 

\subsection{Classifier With a Reject Option}

Multi-class classification refers to classifying observations that
take values in an arbitrary feature space $\mathcal{X}$ into one of
$N_{c}$ classes. A discriminant function $f$ : $\mathcal{X}$
$\rightarrow$ $\mathcal{R}$ yields a classifier $f(x) \in \{1, ...,
N_{c}\}$ that represents the guess of the label $Y$ of a future
observation $\mathcal{X}$ and the error if the $y \neq f(x)$. Bayes decision rule assigns
each pattern $x$ to the class $\omega_i$ for which the a posteriori
probability $P(\omega_{i} \mid x)$ is maximum. 

\begin{equation}
P(\omega_{i} \mid x)= max_{k=1,..., N_{c}} P(\omega_{k} \mid x) 
\end{equation}

Since observations $x$ for which the conditional probability $P(\omega_{i} \mid x)$
is close to 1/2 are difficult to classify, we introduce a reject option for classifiers,
by allowing for a third decision called $\circledR$ (reject),
expressing doubt, to be possible.

The reject option is built by using a threshold value $0 \leq \tau < 1$ as follows.
Given a discriminant function $f$ : $\mathcal{X}$ $\rightarrow$ $\mathcal{R}$, we report $f(x) \in \{1, ..., N_{c}\}$ if
$\mid f(x)\mid > \tau$, but we withhold decision if $\mid f(x)\mid \leq \tau$ and report $\circledR$. 

The classifier's accuracy is defined as the conditional probability that a pattern is correctly
classified, given that it has been accepted:

\begin{equation}
\begin{array}{rl}
Accuracy = P(correct \mid accepted) \\
\\
 = \frac{P(correct)}{P(correct)+P(error)} 
\\
\end{array}
\end{equation}

We assume
that the cost of making a wrong decision is 1 and the cost of utilizing the reject
option is $d > 0$. Value of $d$ includes both cost of passing time and
lack of output. The appropriate risk function is then the following:

\begin{equation}
\begin{array}{rl}
\mathbb{E}[l(Y f(X))] = \mathbb{P} \{ P(\omega_{i} \mid
x_{\omega_{j}}) > \tau \} + \\

d\mathbb{P} \{ P(\omega_{i} \mid x_{\omega_{i}}) \leq \tau \}
\\
\end{array}
\end{equation}

for the discontinuous loss

\begin{equation}
l(z) = \left\{
\begin{array}{rl}
1 & if  P(\omega_{i} \mid x_{\omega_{j}}) > \tau,\\
d & if  P(\omega_{i} \mid x_{\omega_{i}}) \leq \tau,\\
0 & otherwise.
\end{array} \right.
\end{equation}

Since a reject decision is never triggered if $d > 1/2$, see \cite{Herbei2006}, we restrict to the cases
$0 \leq d \leq 1/2$. The generalized Bayes discriminant function, minimizing (2), is
then \cite{Chow1970}, \cite{Ripley1996}:

\begin{equation}
f_{o}(x) = \left\{
\begin{array}{rl}
i & if  P(\omega_{i} \mid x_{\omega_{i}}) > d,\\
0 & if  P(\omega_{i} \mid x_{\omega_{j}}) \leq d,\\
\end{array} \right.
\end{equation}

with risk 

\begin{equation}
\mathbb{E}[min {f(x), 1-f(x), d}]
\end{equation}
 
The case $(\tau, d) = (0, 1/2)$ reduces to the classical situation without the
reject option. We can view $d$ as an upper bound on the conditional probability
of misclassification (given $X$) that is considered tolerable.

\subsection{Proposed Forefront-Nose for early odor classification}

This subsection introduces a real-time E-Nose system called
\emph{Forefront-Nose} for early and reliable recognition of gas
signals. Earliness of the decision comes from the fact that the
proposed Forefront-Nose is using only a small portion of the
time-series odor signal available at a time, rather than buffering the
entire signal in the memory which would mean more waiting time. The
system calculates the reliability score for a candidate label based on
the first observations from the signal. If the score is high enough to
report it to the user, the E-Nose performs the
classification. Otherwise, the next portion of the signal is given to
the system (obviously increasing \emph{"cost"}). This process can continue until a reliable decision is made or the cost of decision is higher than a user-set threshold. 

\begin{figure}
\center
  \includegraphics[width=6cm,height=2cm]{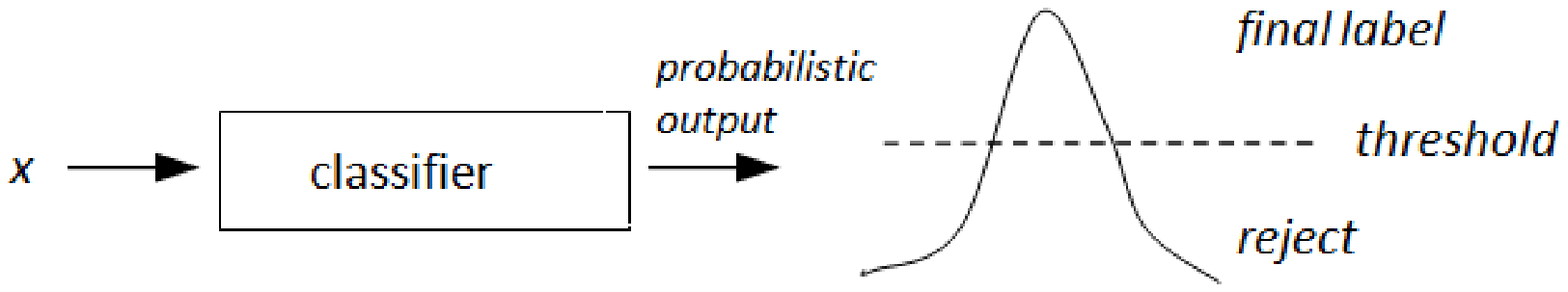}
   \includegraphics[width=6cm,height=3cm]{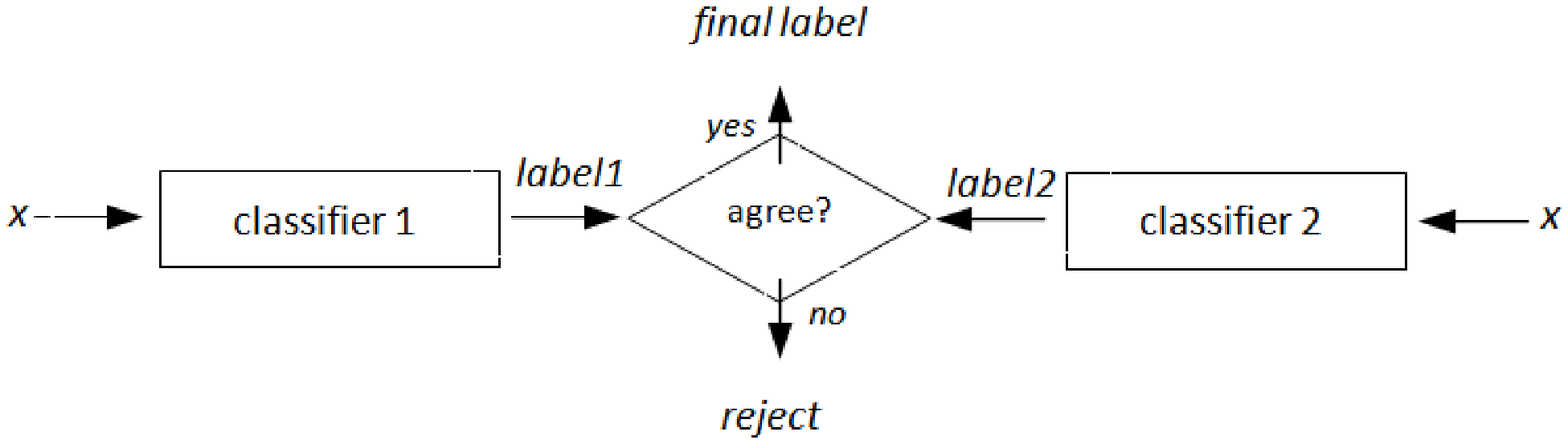}
\caption{Standard CWRO (up) vs. the proposed CWRO (down). Standard
  CWRO are limited only to classifiers with ability to calculate
  posterior probabilities and need a user-set threshold. The proposed
  CWRO, on the other hand, has none of these drawbacks and is more
  reliable/stable because of hiring an ensemble.}
\label{fig:1}       
\end{figure}

Besides the E-Nose application perspective, Forefront-Nose presents an
important contribution to the machine learning and pattern recognition
field. The main core of the proposed Forefront-Nose is a novel
CWRO. Let us recall that the standard approach to build a CWRO hires a
single classifier to calculate the "posterior" probability and then
applies a threshold to reject or accept the decision. There are two
drawbacks of such a system as follows: the first one is the need for
one extra step for tunning the threshold and the other drawback is the
dependency between the classifiers stability and the reliability of
the output, meaning that if something goes wrong with the classifier,
it can lead to a wrong decision by entire system. The proposed novel
CWRO model addresses both these issues by hiring two classifiers
(instead of one single expert) and using their agreement to
accept/reject an output (see Fig. \ref{fig:1}).

Another advantage of such a system is that it is not necessarily
limited to the classifiers with probabilistic outputs. For instance,
standard kNN or SVM algorithms can not be used for CWRO unless further
operations applied first to obtain the posterior
probability. Furthermore, the standard way of taking advantage of an
ensemble idea in CWRO is to combine classifiers (e.g. voting strategy)
in order to obtain the ensemble posterior probability and then apply a
threshold. In the proposed approach, "ensemble consensus" directly
plays a role about accept/reject decision. 

The Forefront-Nose model is
outlined in Algorithm \ref{Forefront-Nose}. 

\begin{algorithm}[htbp]

\vspace{0.2cm}

Set parameters $C=\{ c_{1}, c_{2}, ... c_{i}\}$, $\gamma=\{ \gamma_{1}, \gamma_{2}, ... \gamma_{j}\}$ and $w_{k}$ being $k=1st,2nd, ...$ time interval available in time.

\vspace{2mm}

{\bf TRAINING}

$\hspace{0mm}$for each $k$ do:

$\hspace{5mm}$for each $(c_{i}, \gamma_{j})$ do:

$\hspace{10mm}$Build a classifier $f_{ij}$

$\hspace{5mm}$Select $n$ most accurate classifiers

$\hspace{5mm}$Calculate $DF$ for each pair of the selected classifiers

$\hspace{5mm}$Build $n \times n$ diversity matrix

$\hspace{5mm}$Select the pair $P_{k}$ with highest diversity for CWRO

\vspace{2mm}

{\bf TESTING}

$\hspace{0mm}$for each $k$ do:

$\hspace{5mm}$Apply $P_{k}$ to make accept/reject decision based on the agreement rule

$\hspace{5mm}$If decision=accept do:

$\hspace{10mm}$ $final\_label$= $given\_label$

$\hspace{10mm}$ exit

\caption{Forefront-Nose for early odor classification. This algorithm is typically written for SVMs, however, it can be adapted to any classifier type. \label{Forefront-Nose}}
\end{algorithm}

There is no agreement under what circumstances an ensemble of
classifiers obtains a better accuracy compared to the
individuals. However, it is both practically and theoretically proven
that an ensemble of classifiers is more stable (reliable) in the
decision making process. There are two main factors to be considered
to build a \emph{"good"} ensemble: individual accuracy and diversity
among them. According to this idea, several studies advocate the
method of producing a pool of classifiers followed by a selection
procedure to pick the classifiers that are most diverse and accurate
\cite{Kuncheva2004}. In Forefront-Nose, we first build an SVM
classifier for each $(C,\gamma)$ (the SVM-specific parameters) and select $n$ classifiers with best
recognition rates. The diversity measure is then used to form a
pairwise diversity matrix for a classifier pool and subsequently to
select classifiers that are least related. For this purpose, we use
the double fault measure \cite{Kuncheva2004} in this study. The double fault measure is an intuitive choice for diversity measure as it gives the probability of both classifiers $f_{i}$ and $f_{j}$ being incorrect:

\begin{equation}
DF_{i,j}=e
\end{equation}

where $e$ is the ratio of the number of samples which both classifiers misclassified at the same time to total number of evaluated samples. This measure is based on the concept that it is more important to know when simultaneous errors are
committed than when both classifiers are correct. Thus this measure is
appropiate for incorporation in the proposed CWRO. It is worth noting that $DF$ is a \emph{pairwise} measure that considers a pair of classifiers at the time. Therefore, the diversity matrix for $n$ given classifiers is a $n \times n$ symmetric matrix with diagonal array representing minimum pairwise diversity value.

\section{Experimental Results}
\label{sec:8}

Computational experiments focus on an extensive real data set collected in the wind tunnel test-bed
facility \cite{Vembu2012}. In the following subsections, we will first describe the
sensors used in our experiments as well as the data sets and the
measurement protocol. We will then present
the experimental results obtained from applying the proposed Forefront-Nose on the wind tunnel dataset.

\subsection{Forefront-Nose Hardware Description}

\begin{figure}
\center
  \includegraphics[width=6cm,height=2.5cm]{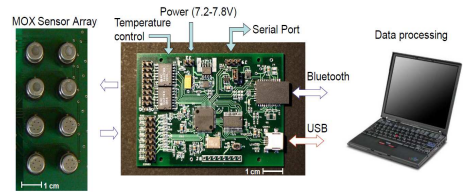}
\caption{The custom-designed portable tin-oxide gas sensor array used as the sensing layer of the chemical sensory system (from \cite{Vembu2012}).}
\label{fig:2}       
\end{figure}

Chemo-resistive sensing principles, such as polymers and metal-oxide based sensor technologies,
are sensitive to rapid changes in the analyte concentration at the measurement location, while performing reasonably well in discriminating the said chemical analytes \cite{Carmel2003}. Their high spatio-temporal resolution distinguishes them as suitable receptors
in ambient conditions, especially for the prediction problems addressed in this paper
(i.e., gas discrimination/identification and gas source localization). Accordingly, we utilized
a portable array module endowed with 8 commercialized metal-oxide gas sensors, provided
by Figaro Inc. \cite{figaro}, to record a gas plume generated in the wind tunnel facility (see Figure \ref{fig:2}). The sensing principle is based on oxidation/de-oxidation of the analyte while
in interaction with the active surface. Such oxygen/electron exchange between the pollutant
analyte and the sensitive surface induces a change in the surface conductance, which
is measured in the form of a resistance, or conductance, time-series across the electrodes of
the sensor \cite{Pearce2003}. 

In our particular experiment, we set the operating temperatures by following an empirical
analysis described as follows: (i) In a gas delivery manifold (i.e., an isolated chamber series
to a computer-supervised continuous 
flow system), each sensor was repeatedly subjected to
each candidate analyte by increasing concentration. (ii) Then, for each sensor type, the
admissible range of surface temperatures (i.e., sensor's heater voltages) was swept, and the
value maximizing the response dynamic range over these presentations was determined. (iii)
This empirical optimum is assigned as the measurement parameter for just one of the two
sensors of that specific sensor type on the portable array. The heater voltage of the other
element in the pair is then set to 5 Volts2, which is a generic value equal to the mid-point
of the admissible voltage range suggested by the manufacturer.

\subsection{Dataset description}

A gas plume emitted from a fixed chemical source location conveys two critical pieces
of information to the spatial coordinates within its volume: the analyte identity and the
displacement vector from the source to the observer. The problems of chemical source identification and localization are individually not novel to the fields of sensory signal processing
and pattern recognition. While chemical identification is a genuine classification problem,
in which various mainstream methods have been successful thanks to the executive quantifications and metrics that have been recently established in the chemosensory community \cite{Hierlemann2008}.

\begin{figure}
\center
\includegraphics[width=5cm,height=2cm]{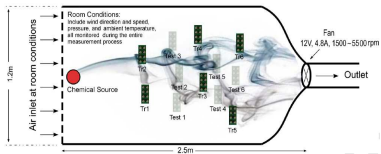}
\caption{Wind tunnel test bed facility used to collect time series data from sensor arrays for the problem
of gas identification (from \cite{Vembu2012}).}
\label{fig:3}       
\end{figure}

We collected the data set utilizing a portable sensor array module endowed with eight
metal-oxide gas sensors manufactured by Figaro Inc. \cite{figaro} positioned at different fixed locations
within a gas plume generated in the wind tunnel test-bed facility. The first data set
induces a ten-gas classification problem, in which the goal is to discriminate four analytes,
namely, carbon monoxide, ammonia, methane, acetaldehyde, benzene, butanol, ethylene, methanol, toluene and acetone, regardless of the
location of the sensory system module within the annotated wind tunnel test-bed facility.
It comprises 504 8-dimensional time series responses each dimension corresponds to each sensor utilized observed at the locations (that we call landmarks) indicated in the wind tunnel shown in
Figure \ref{fig:3}. This data set thus induces a ten-gas
classification problem, i.e., determining the odor chemical source identity regardless of the
position of the sensing layer module.

\subsection{Data acquisition and measurement procedure}

In collecting our above described data set, we utilized a 2.5m $\times$ 1.2m $\times$ 0.4 m wind tunnel,
where the geometry of the problem, the location of the odor source and our chemo-sensory
platform, as well as the controllable flow conditions are shown, for each case of study, in Figure \ref{fig:3}. The considered wind tunnel has a non-inclined floor, which allows
us to disregard the altitude dimension of the field. In constructing the above mentioned data
sets, we adopted the following protocol. First, an artificial air flow of 0.10 m/s was induced
into the wind tunnel by an exhaust fan rotating at a constant speed of 1500 rpm, which,
in turn, allowed us to, both, avoid saturation of the test volume to analytes and provide a
turbulent flow in the wind tunnel, as reflected in the time series responses recorded from the
chemical sensors under the turbulent environment (see Figure \ref{fig:2}). The air
flow was measured utilizing two 2-D Ultrasonic Anemometers, by Gill Windsonic, each of which allocated
at different positions throughout the entire wind tunnel before starting the recordings of the
analytes. Afterwards, one of the above pre-defined fixed locations was selected to allocate
our chemo-sensory module and the wind tunnel was then sealed to avoid the presence of leaks
and external interfering odors in the room and wind tunnel, respectively, while performing
our measurement recordings. Once the air
flow and the remaining ambient conditions have
assume a quasi-stationary situation in the wind tunnel, one of the analytes was randomly chosen and
released at the source location mark, indicated in the figure \ref{fig:3}, and held circulating
through the tunnel for several minutes before starting the actual recordings. This stage
constitutes a preliminary phase, which was utilized to "stabilize" the sensors' responses in
the presence of the chemical anlaytes in the wind tunnel and ensure that only the stochasticity
of the turbulent air
flow in presence of the different substances being released in the
wind tunnel were reflected in the sensors' responses.

Subsequently, each measurement, that consisted of
an 8-channel time series representing the time profile of the sensor resistances in response to
each analyte being studied, was recorded for three minutes and stored in our data set repository
for further processing. After that step, the source was removed, the wind tunnel was
ventilated and the hood was left open for one minute, before a new measurement location
and/or analyte type were set for subsequent measurements. This measurement procedure
was exactly recreated for each chemical analyte and landmark location until all pairs were covered.
Each measurement was recorded at a sampling rate of 100 Hz. Note that there
is no symmetry in the spatial distribution of a plume with respect to the main axis (i.e.,
the line connecting the source to the exhaust). A plume demonstrating a perfect symmetry
in real conditions is rare due to the non-symmetry of the volume enclosing the field, the
inhomogeneous temperature, and the variability of the 
ow direction. In the setup phase of
the data collection we found in all trials that the selected features evaluated at symmetric
coordinates were strictly different. Yet, as we show next, the non-symmetric plume structure
is reproducible so that the predictions can be extended in time, i.e., by referring to earlier
observations at the same environment.

In all our experiments, we reduced the length of time series (i.e., down-sampled) by
averaging over windows of size 10 samples within the same time series in order to speed up
computations. 

\subsection{Numerical results and comparisons}

To evaluate the proposed forefront-nose, the classification
performance assessed by the 10-fold cross-validation provides
realistic generalization accuracy for unseen data. Also, we calculated the deviation of a time-series data and set a threshold to detect the rising point of the signal i.e. presence of a gas in the wind tunnel. In all experiments, we have used SVM classifiers with a RBF kernels, varying both $\gamma$ and $C$ from $2^{-5}$ to $2^{5}$ in order to find a best parameters. 

\begin{figure}
\center
\includegraphics[width=8cm,height=5cm]{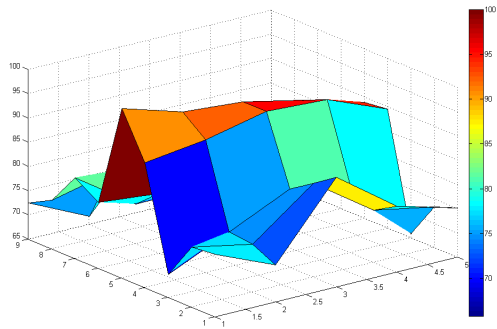}
\caption{Accuracy of a standard classifier on the wind tunnel test-bed facility. The surface indirectly shows the locations where the signal is more strong and discriminable. }
\label{fig:4}       
\end{figure}

\begin{figure}
\center
\includegraphics[width=8cm,height=5cm]{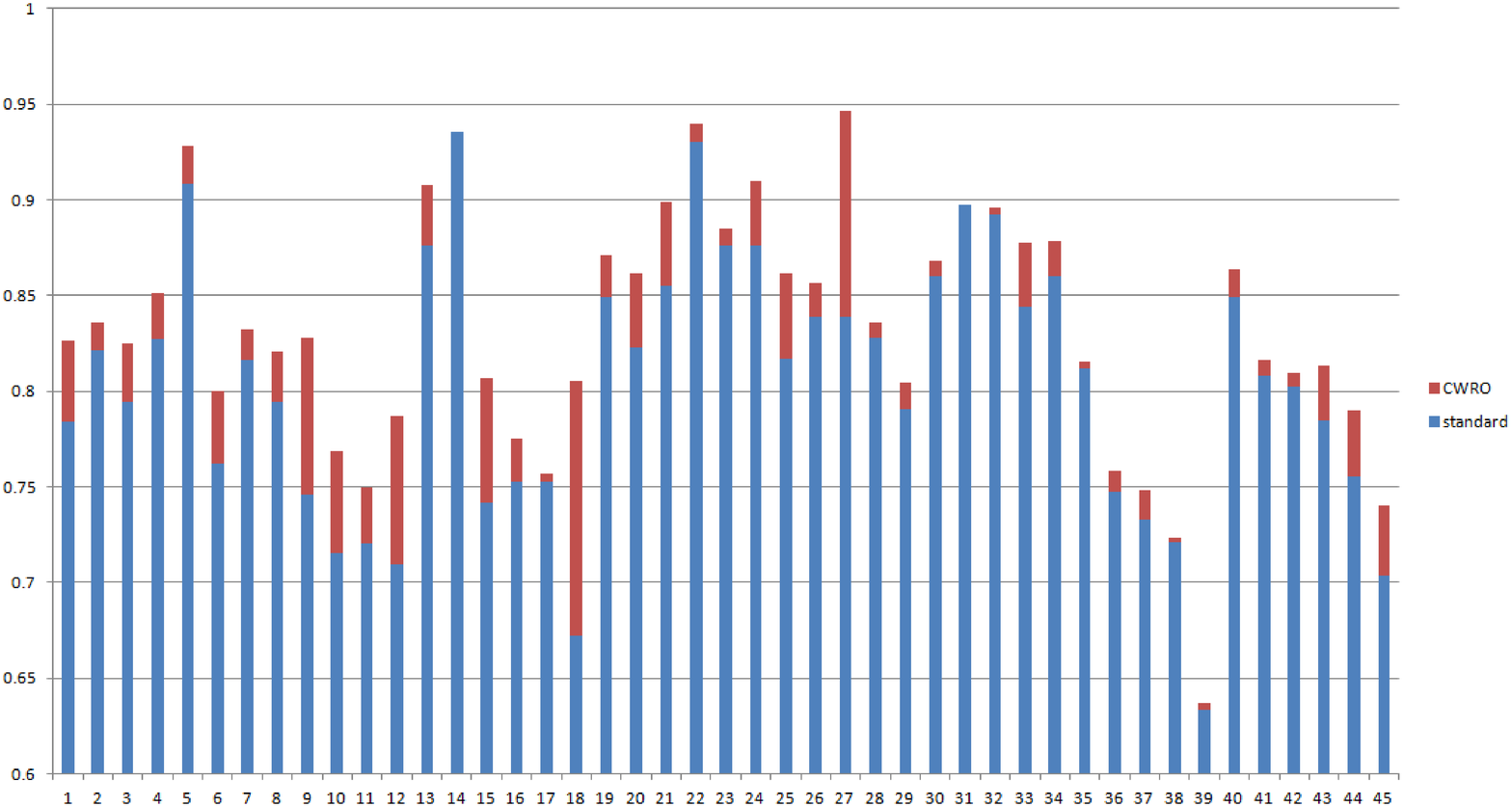}
\caption{Early recognition of the classifier with a reject option versus the standard classifier of the same kind (SVM with RBF kernel, in this case).}
\label{fig:5}       
\end{figure}

In the first phase of our experiments, we have run a classifier on all
over the wind tunnel to investigate the signal strength and
discriminability (see Figure \ref{fig:4}). Then, we have compared the
recognition rate of a standard classifier with a reject option to the
standard classifier to see possible drop/raise in the
performance. First $k$ points of a time-series signal is given as an
input, since we are interested in early recognition. In this
particular case, we choosed $k=50$ which is equivalent of
5 seconds. The results are shown in Figure \ref{fig:5} in which the x
axis indicates the index of the 45 locations spread all over the wind
tunnel. As shown, the results encourage the use the CWRO which has a
performance superior to a standard classifier. 

Furthermore, in order to verify the improvement related to early
recognition, we compared the performance of the proposed
forefront-nose with the standard CWRO (see Table \ref{tab:9}). The comparison is made for
different earliness of a time-series in order to obtain results with
no dependency with the signal length. Therefore, we have varied $k$
from 5 to 30 seconds. We report the results for each $k$ by averaging
the recognition rate over all 45 locations in the wind
tunnel. The comparative results confirm the robustness of the
proposed forefront-nose with respect to the commonly used standard CWRO. Since the forefront-nose gets the maximum performance using $k=10s$, this amount of time seems to be optimum for both early and accurate recognition of different gas types.

\begin{table}[t]
\caption{Average recognition rates of the forefront-nose compared to CWRO over different locations of the wind tunnel.}
\label{tab:9}       
\begin{tabular}{lllllll}
\hline\noalign{\smallskip}
earliness  & 5s & 10s & 15s & 20s & 25s & 30s  \\

\noalign{\smallskip}\hline\noalign{\smallskip}

Std. CWRO & 90.15 & 91.99 & 91.68 & 90.93 & 88.5 & 87.4 \\

Forefront-Nose  & 93.20 & 94.10 & 93.14 & 93.50 & 91.04 & 90.1 \\
\noalign{\smallskip}\hline
\end{tabular}
\end{table}

\section{Conclusions and future work}

A new architecture is proposed for a classifier with a reject option
using ensemble of experts. The approach requires the ensemble's
"agreement" in order to accept a candidate label rather than relying
only on single expert's posterior probability as suggested by standard
methods. The proposed algorithm is engaged as a core mechanism to
address the problem of odor/gas discrimination in an online E-Nose
called \emph{Forefront-Nose}. The method forces the decision making as
early as possible and is shown to obtain a high accuracy on a real dataset. 

Future work focuses on two main issues: source localization i.e. automatic detection and recognition of an odor/gas source, and the "generalization" which is the problem of classifying a gas type regardless of its location on x-y axis.

\section*{Acknowledgment}
Camelia Chira acknowledges the support of Grant PN II TE 320, Emergence, auto-organization and evolution: New computational models in the study of complex systems, funded by CNCS Romania.



%

\end{document}